\documentclass[runningheads]{llncs}
\usepackage[T1]{fontenc}
\usepackage{graphicx}
\usepackage{booktabs}
\usepackage{amsmath}
\usepackage{amssymb}
\usepackage[misc]{ifsym}
\usepackage[numbers]{natbib}

\usepackage{booktabs}
\usepackage{multirow}
\usepackage{enumitem}
\usepackage{longtable}
\newcommand{\corr}{(\Letter)}

\newlist{inlinedesc}{description*}{1}
\setlist[inlinedesc]{itemjoin={{ $\odot$ }}, itemjoin*={{ and }}}

\begin{document}

\title{Demographically-Informed Heat-Mortality Risk Curves via Risk Graph Neural Networks}

\titlerunning{Heat-Mortality Risk with GNNs}

\author{Alex O. Davies\inst{1} \corr \and
Eunice Lo\inst{1} \and
Rui Zhu\inst{1}}


\authorrunning{Davies et al.}

\institute{School of Geographical Sciences, University of Bristol, UK
\email{alexander.davies@bristol.ac.uk}}

\maketitle

\begin{abstract}

Estimating heat-related mortality risk is a core task in environmental epidemiology, typically addressed with Distributed Lag Non-linear Models (DLNMs); interpretable exposure-response surfaces fitted to temperature-mortality time series.
DLNMs are effective but ignore demographic and geographic context, despite well-established relevance to heat vulnerability. 
We propose Risk Graph Neural Networks (RGNNs), a hierarchical GNN encoder that uses granular census features to optimise DLNM coefficient vectors, preserving interpretable risk curve outputs while substantially improving predictive calibration.
Evaluated across 10 regions of England and Wales on two unprecedented heat years, RGNN variants maintain both lower point-errors and near-nominal uncertainty coverage during the 2022 heatwave where baselines collapse.
\keywords{temperature--mortality modelling 
\and graph neural networks
\and distributed lag non-linear models
\and spatial epidemiology}
\end{abstract}

\section{Introduction}

A central task in environmental epidemiology is estimating mortality risk from temperature exposure. 
The dominant tool is the Distributed Lag Non-linear Model (DLNM) \cite{gasparriniDistributedLagNonlinear2010}, which produces an interpretable exposure-response surface over a lag dimension.
This captures, for instance, that cold mortality peaks 7–14 days after exposure while heat mortality concentrates in the immediate 3-day window. 
DLNMs operate purely on temporal signals, however, making no use of demographic or geographic features despite their established importance in mediating heat vulnerability. 
Recent ML approaches \cite{boudreaultMultiregionModelsBuilt2024} sometimes improve point estimates but rarely match the interpretability of the DLNM risk curve, and \citet{boudreaultMachineLearningModelling2025} note that the DLNM is rarely used as a proper benchmark.

A core reason for the wide adoption of DLNMs over other predictive models is the interpretability of their outputs.
A risk curve is immediately transparent to technical and non-technical users, making it a powerful tool in informing policy decisions.
We introduce Risk Graph Neural Networks (RGNNs), which augment DLNMs with a hierarchical GNN encoder over census features. 
The result preserves the interpretable risk curve while substantially improving calibration under distributional shift — precisely the regime that matters most for public health decision-making during extreme heat events. 
Early counterfactual experiments produce spatially-resolved, policy-relevant mortality predictions across urban areas.

\section{Background}

\begin{figure}[h]
    \centering
    \includegraphics[width=\linewidth]{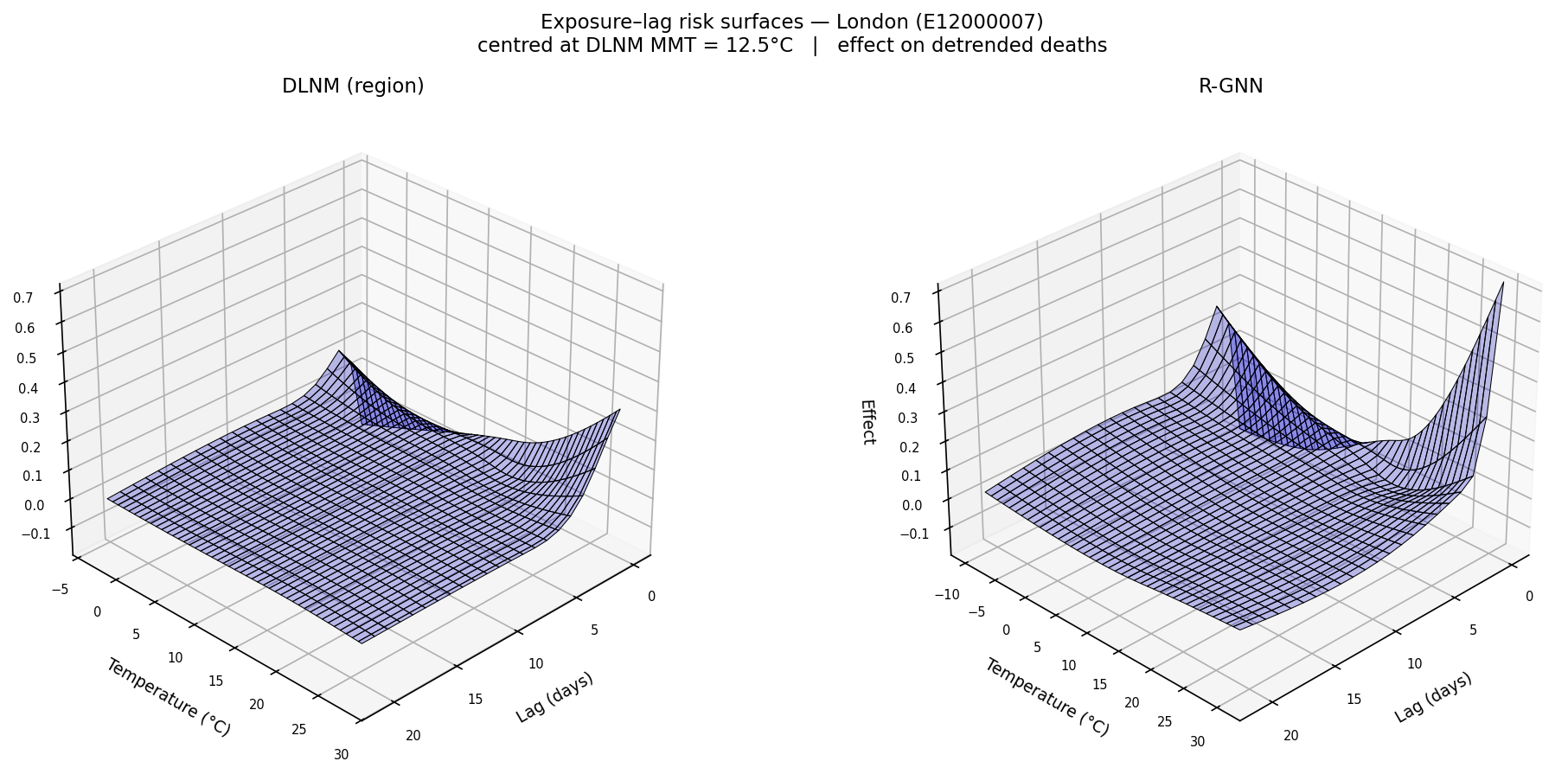}
    \caption{Risk surfaces, in units of deaths per 100k, for London. Left: Vanilla DLNM. Right: RGNN-Calibrated.}
    \label{fig:risk-surfaces}
\end{figure}

Here we briefly summarise the current state of modelling the heat-mortality relationship.
DLNMs are proposed and formalised in-detail in \citet{gasparriniDistributedLagNonlinear2010}, and \citet{boudreaultMachineLearningModelling2025} review ML approaches.

\subsection{Distributed Lag Non-linear Models (DLNMs)}

A DLNM~\cite{gasparriniDistributedLagNonlinear2010} models a scalar health outcome at location $i$ and time $t$ as a linear combination of a \textit{cross-basis} matrix that encodes the joint non-linear and lagged relationship between an exposure and the outcome:

\begin{equation}
    \hat{y}_{it} = \boldsymbol{w}_i \cdot \mathbf{CB}_{it},
\end{equation}

where $\boldsymbol{w}_i \in \mathbb{R}^{n_{cb}}$ is a coefficient vector estimated by ordinary least squares and $\mathbf{CB}_{it} \in \mathbb{R}^{n_{cb}}$ is the cross-basis row for location $i$ at time $t$.
The primary output is a exposure-outcome effect surface along lag and the exposure variable, which in our case is lag-temperature-risk, as visualised in Figure~\ref{fig:risk-surfaces}.
Fitting one DLNM per location independently produces unstable estimates for small areas with sparse death counts, and provides no mechanism for borrowing strength from neighbouring or demographically similar locations.

\subsection{Graph Neural Networks (GNNs)}

Graph Neural Networks~\cite{kipfSemiSupervisedClassificationGraph2017} generalise neural networks to graph-structured data by iteratively aggregating information from local neighbourhoods.
GATv2~\cite{velickovicGraphAttentionNetworks2018} extends graph attention networks with a more expressive attention mechanism in which both source and target node features jointly determine attention weights, making it well-suited to heterogeneous hierarchies where child nodes vary substantially in informativeness.
GraphSAGE~\cite{hamiltonRepresentationLearningGraphs2017} performs mean aggregation over sampled neighbourhoods and is widely used for spatial smoothing on geographic graphs.
GNNs have been applied to urban analytics tasks including traffic forecasting~\cite{jiangGraphNeuralNetwork2022} and air quality prediction~\cite{fillolaEnablingFastGreenhouse2026}, but their application to epidemiological exposure-response estimation is, to our knowledge, novel.

\section{Risk Graph Neural Networks (RGNNs)}
 
\begin{figure}[h]
    \centering
    \includegraphics[width=\linewidth]{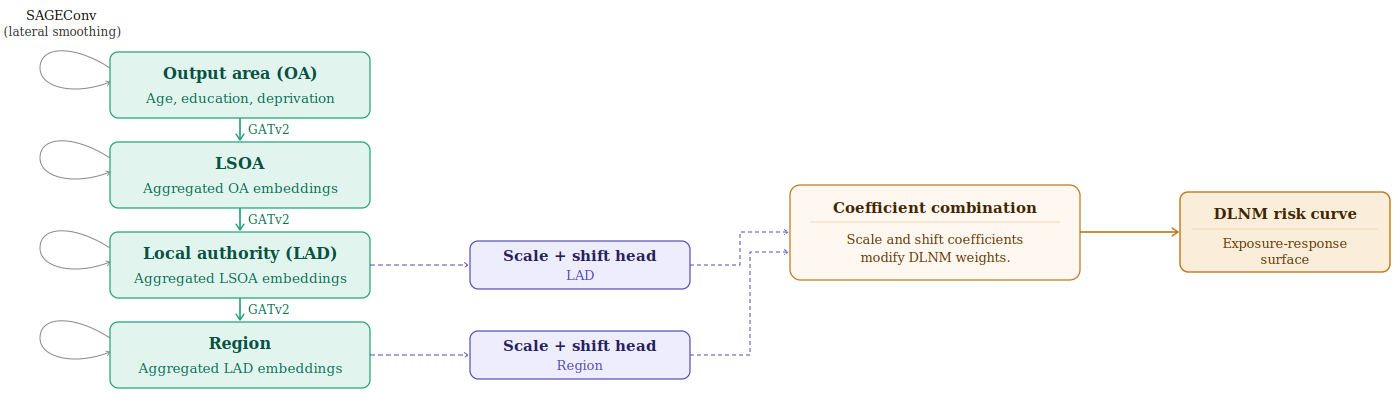}
    \caption{RGNN architecture. Census features at the lowest
    administrative level are projected to a hidden
    embedding and propagated upward and laterally in separate stages, with heads predict modifying weights (Eqn.~\ref{eqn:rgnn}) over an OLS-initialised DLNM.}
    \label{fig:arch}
\end{figure}

The RGNN replaces per-location independent OLS with a learned mapping from census features to DLNM coefficient vectors, structured as a two-level hierarchy: a frozen per-region OLS baseline is modulated by GNN-derived corrections at both the region and LAD levels.
The output preserves both the interpretability and strong prior of the OLS-initialised DLNM, while improving performance with spatially-shared demographic information, as well as being comparably cheap to run compared to full ML models.
An overview of the architecture is given in Figure~\ref{fig:arch}.
 
\paragraph{Initialisation.}
Before GNN training, per-region OLS coefficients $\boldsymbol{b}_r$
are estimated on region-level Pearson mortality residuals and frozen. 
All GNN heads are zero-initialised, so at the start of training every LAD's coefficient vector equals its region OLS baseline.
 
\paragraph{Encoder.}
A bottom-up heterogeneous GNN propagates static demographic features from the finest available administrative unit upward through the hierarchy. 
At each level, child embeddings are aggregated into parent nodes using multi-head \emph{GATv2Conv} attention; geographic neighbours at the same level are smoothed using \emph{SAGEConv} mean aggregation. 
Higher-level nodes receive no direct demographic features; their representations are built entirely from upward message passing, forcing the model to learn meaningful aggregations.
 
\paragraph{Coefficient prediction.}
Zero-initialised linear heads at each predictive level map GNN embeddings to corrections in $\mathbb{R}^{n_{cb}}$:
\begin{equation}
\label{eqn:rgnn}
    \boldsymbol{w}_i
    = \boldsymbol{b}_{r(i)}
    + \mathrm{clamp}\!\left(
        \boldsymbol{b}_{r(i)} \odot f_{\mathrm{ls}}(\mathbf{h}_i)
        + f_{\mathrm{lsh}}(\mathbf{h}_i),\;
        \pm\delta
      \right),
\end{equation}
where $\odot$ is element-wise multiplication, $\delta$ is a hyperparameter controlling the maximum correction magnitude, and the scale and shift heads modulate the shape and location of the baseline curve respectively.
Importantly, weights at each level anchor independently to the same frozen baseline rather than being chained, keeping granularities independently interpretable and preventing cascading errors.
With dropout, and repeat predictions of coefficients, we can form useful uncertainties that can be propagated into the vanilla DLNMs uncertainty estimations.
 
\paragraph{Training objective.}
The primary loss is Lin's CCC \cite{linConcordanceCorrelationCoefficient1989}, which jointly penalises errors in correlation, variance ratio, and mean bias, summed at LAD and region levels with population weighting. Three auxiliary terms are added: (i) a monotonicity penalty encouraging physiologically plausible cold and heat tails; (ii) a consistency penalty penalising divergence between population-weighted LAD aggregates and region-level predictions; and (iii) a population-inverse-weighted L2 delta penalty pulling data-sparse LADs toward their regional baseline. MC-dropout over the coefficient heads propagates uncertainty into the final risk curves.

\section{Experiments}

\begin{table}[t]
\centering
\caption{Region-level holdout evaluation across 10 regions of England and Wales.
Values are mean $\pm$ standard deviation across regions. Bold indicates the best result
in each column.}
\label{tab:region_holdout}
\small
\begin{tabular}{llccccc}
\toprule
\cmidrule(lr){3-7}
Year & Model & RMSE $\downarrow$ & MAE $\downarrow$ & Pearson $\uparrow$ & CRPS $\downarrow$ & Cov@90\% $\uparrow$ \\
\midrule
\multirow{8}{*}{2018} & DLNM & 0.33 $\pm$ 0.07 & 0.25 $\pm$ 0.05 & 0.32 $\pm$ 0.05 & 0.19 $\pm$ 0.04 & 0.73 $\pm$ 0.03 \\
 & RF & 0.29 $\pm$ 0.06 & 0.22 $\pm$ 0.04 & 0.45 $\pm$ 0.04 & 0.19 $\pm$ 0.04 & 0.36 $\pm$ 0.02 \\
 & GBM & 0.28 $\pm$ 0.06 & 0.21 $\pm$ 0.04 & 0.37 $\pm$ 0.06 & 0.16 $\pm$ 0.03 & 0.78 $\pm$ 0.03 \\
 & LSTM$^\dagger$ & 0.27 $\pm$ 0.04 & 0.21 $\pm$ 0.03 & \textbf{0.55 $\pm$ 0.10} & --- & --- \\
 & MLP$^\dagger$ & 0.29 $\pm$ 0.05 & 0.23 $\pm$ 0.04 & 0.48 $\pm$ 0.07 & --- & --- \\ \cmidrule(lr){3-7}
 & RGNN & \textbf{0.24 $\pm$ 0.04} & \textbf{0.19 $\pm$ 0.03} & 0.52 $\pm$ 0.08 & \textbf{0.14 $\pm$ 0.02} & 0.97 $\pm$ 0.02 \\
 & RGNN-LSOA & \textbf{0.24 $\pm$ 0.05} & \textbf{0.19 $\pm$ 0.04} & \textbf{0.56 $\pm$ 0.05} & \textbf{0.14 $\pm$ 0.02} & 0.99 $\pm$ 0.01 \\
 & RGNN-OA & \textbf{0.23 $\pm$ 0.04} & \textbf{0.19 $\pm$ 0.03} & \textbf{0.56 $\pm$ 0.05} & \textbf{0.14 $\pm$ 0.02} & 0.98 $\pm$ 0.01 \\
\midrule

\multirow{8}{*}{2022} & DLNM & 0.54 $\pm$ 0.09 & 0.44 $\pm$ 0.07 & \textbf{0.49 $\pm$ 0.11} & 0.36 $\pm$ 0.05 & 0.43 $\pm$ 0.06 \\
 & RF & 0.52 $\pm$ 0.08 & 0.43 $\pm$ 0.06 & 0.46 $\pm$ 0.07 & 0.40 $\pm$ 0.06 & 0.11 $\pm$ 0.03 \\
 & GBM & 0.52 $\pm$ 0.08 & 0.43 $\pm$ 0.06 & 0.40 $\pm$ 0.08 & 0.35 $\pm$ 0.05 & 0.39 $\pm$ 0.07 \\
 & LSTM$^\dagger$ & 0.53 $\pm$ 0.07 & 0.46 $\pm$ 0.06 & 0.40 $\pm$ 0.13 & --- & --- \\
 & MLP$^\dagger$ & 0.53 $\pm$ 0.08 & 0.46 $\pm$ 0.07 & 0.48 $\pm$ 0.11 & --- & --- \\ \cmidrule(lr){3-7}
 & RGNN & 0.39 $\pm$ 0.04 & \textbf{0.31 $\pm$ 0.03} & 0.47 $\pm$ 0.07 & \textbf{0.22 $\pm$ 0.02} & 0.87 $\pm$ 0.03 \\
 & RGNN-LSOA & \textbf{0.39 $\pm$ 0.03} & \textbf{0.32 $\pm$ 0.02} & 0.40 $\pm$ 0.07 & \textbf{0.22 $\pm$ 0.02} & 0.89 $\pm$ 0.03 \\
 & RGNN-OA & 0.43 $\pm$ 0.04 & 0.36 $\pm$ 0.03 & 0.40 $\pm$ 0.07 & 0.25 $\pm$ 0.02 & 0.83 $\pm$ 0.04 \\
\bottomrule
\end{tabular}
\end{table}

We collect daily temperature data from HadUK \cite{hollisHadUKGridANewUK2019} and LAD-level daily mortality through a data-sharing agreement with the UK ONS (2000–2024). For demographic features we use OA-level census information from \citet{goodwinUnifiedDatasetUK2026} for 2021, with the implicit assumption that demographic drift is minimal across the study period. [Adjacent\_To] edges connect areas sharing a boundary, e.g. (Wales) $\leftrightarrow$ (South West); [Contains] edges encode the administrative hierarchy, e.g. (Wales) $\rightarrow$ (Cardiff).

\begin{figure}[h]
    \centering
    \includegraphics[width=\linewidth]{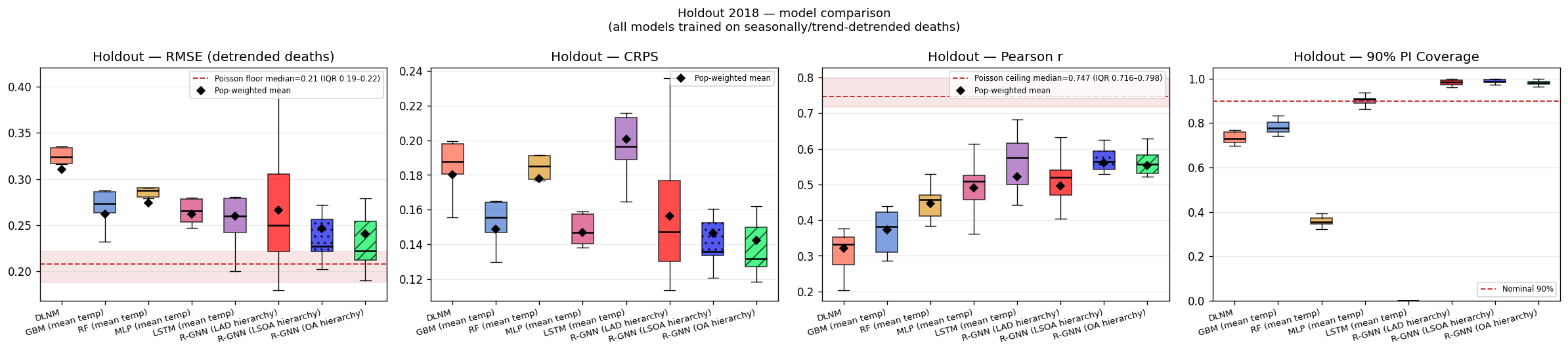}
    \includegraphics[width=\linewidth]{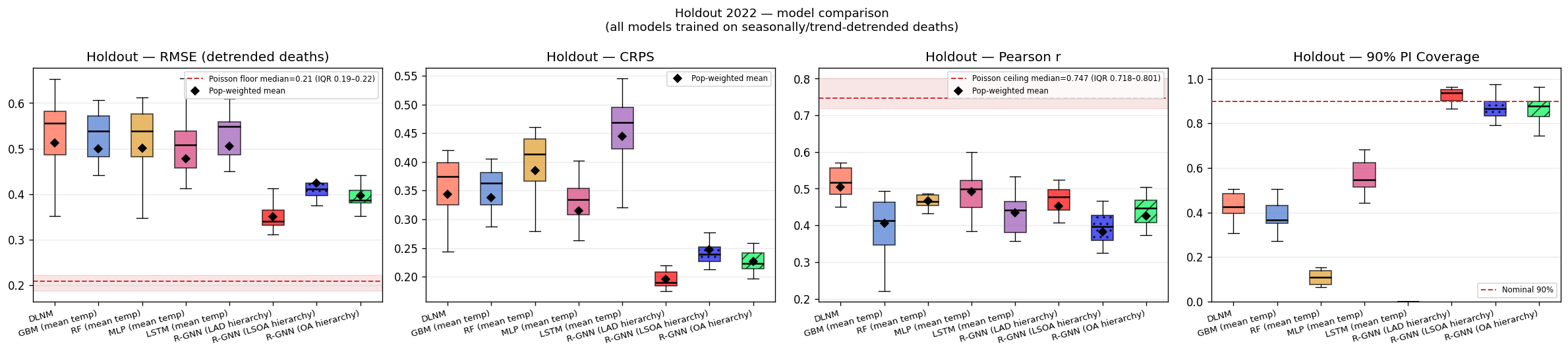}
    \caption{RMSE ($\downarrow$), CRPS ($\downarrow$), Correlation ($\uparrow$) and Coverage@90\% for the holdout years 2018 (top) and 2022 (bottom). }
    \label{fig:placeholder}
\end{figure}

We demonstrate three RGNN variants, which share the same codebase and architecture, differing only in the depth of the administrative hierarchy: 
\textbf{rgnn} (LAD$\to$Region, 2 levels)
\textbf{rgnn\_lsoa} (LSOA$\to$LAD$\to$Region, 3 levels)
\textbf{rgnn\_oa} (OA$\to$LSOA$\to$LAD$\to$Region, 4 levels)
As baselines we use the DLNM, Random Forests (RF), Gradient Boosted Models (GBM), Long Short-Term Memory networks (LSTMs) and Multi-Layer Perceptrons (MLP).
RFs report their own uncertainty estimates through their ensemble, and for GBMs we use heteroscedastic noise on training residuals (std$(y - \hat{y})$).

We take two holdout sets in 2018 and 2022, both unprecedented years for heat exposure in the UK.
We detrend for long-term trend, seasonality, day-of-week effects, and public holidays using linear and Fourier terms.
The outcome $y_{it}$ is a Pearson mortality residual; these are considered our `excess deaths' not covered by these simple detrending terms, including heat-related mortality.
We then normalise units to excess deaths per 100k population.
For each holdout year and model we report RMSE, MAE and Pearson-correlation, as well as CRPS and Coverage@90\% where uncertainty is available.
Hyper-parameters for each model are optimised through a Bayesian search over RMSE with 2017 as validation year, though as is standard practise in environmental epidemiology we use the DLNM parameters from \citet{loEstimatingHeatrelatedMortality2022}, a government-cited study on heat risk in the UK.



\begin{figure}[hbtp]
    \centering
    \includegraphics[width=0.825\linewidth]{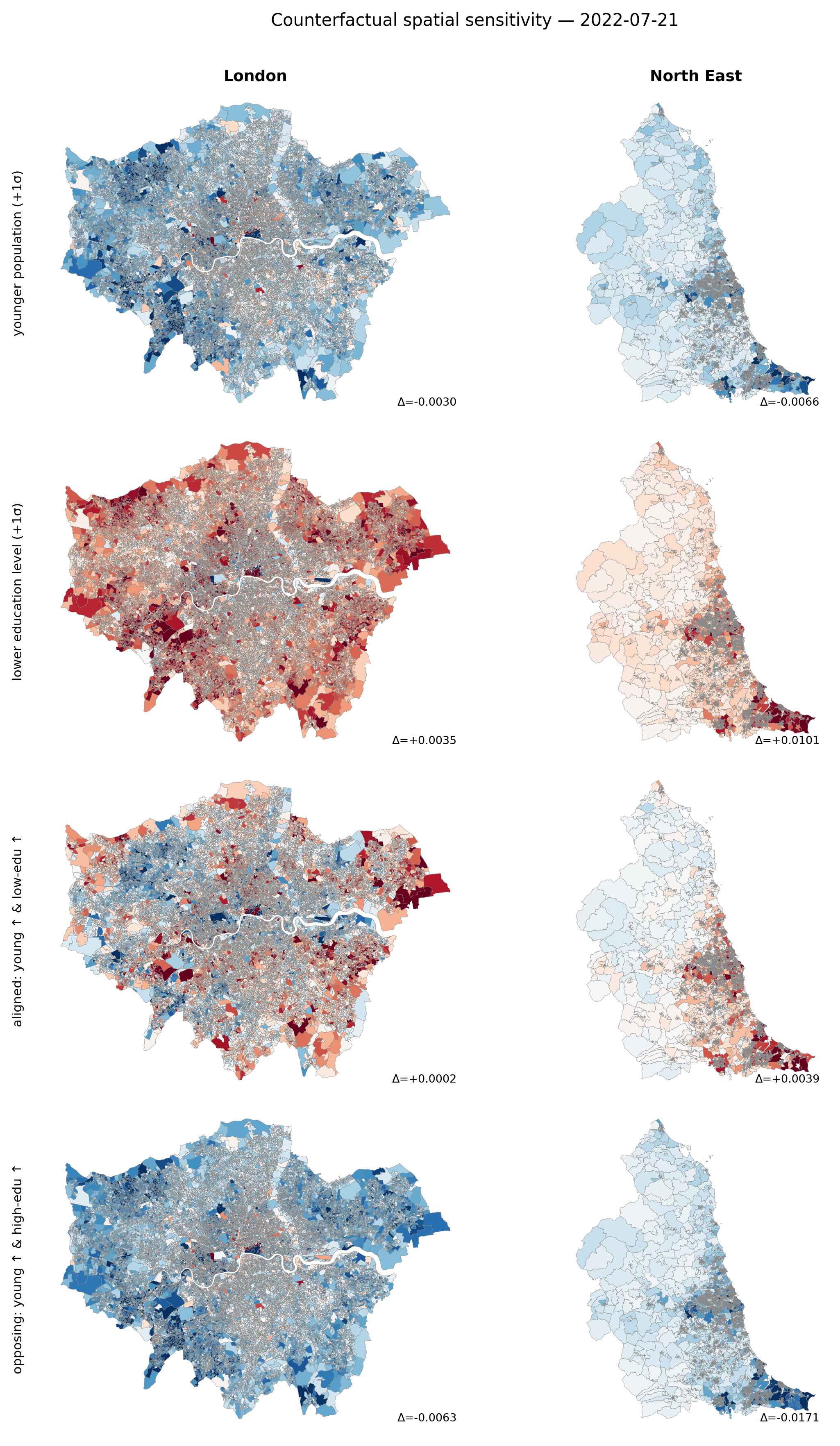}
    \caption{$\Delta$ mortality with varying census features for the RGNN-OA model on the 21st July 2022 in the London and North East regions.}
    \label{fig:counterfactual}
\end{figure}

All three RGNN variants substantially outperform DLNM on RMSE in both holdout years.
In the Supplementary we provide per-region results and example risk curves from the vanilla DLNM alongside the optimised curves from our RGNNs.
The most striking result is the behaviour of predictive coverage
under distributional shift.
In the moderate 2018 holdout, all models with uncertainty estimates
achieve reasonable coverage.
In 2022, however, the DLNM's coverage collapses to well below the nominal 90\% level---while the RGNN
variants maintain coverage between 0.83 and 0.89.
For a public health tool intended to support decision-making during
extreme heat events, calibration under tail conditions is arguably
more consequential than average-year point accuracy.


\subsection{Counterfactuals}

\begin{figure}[h]
    \centering
    \includegraphics[width=\linewidth]{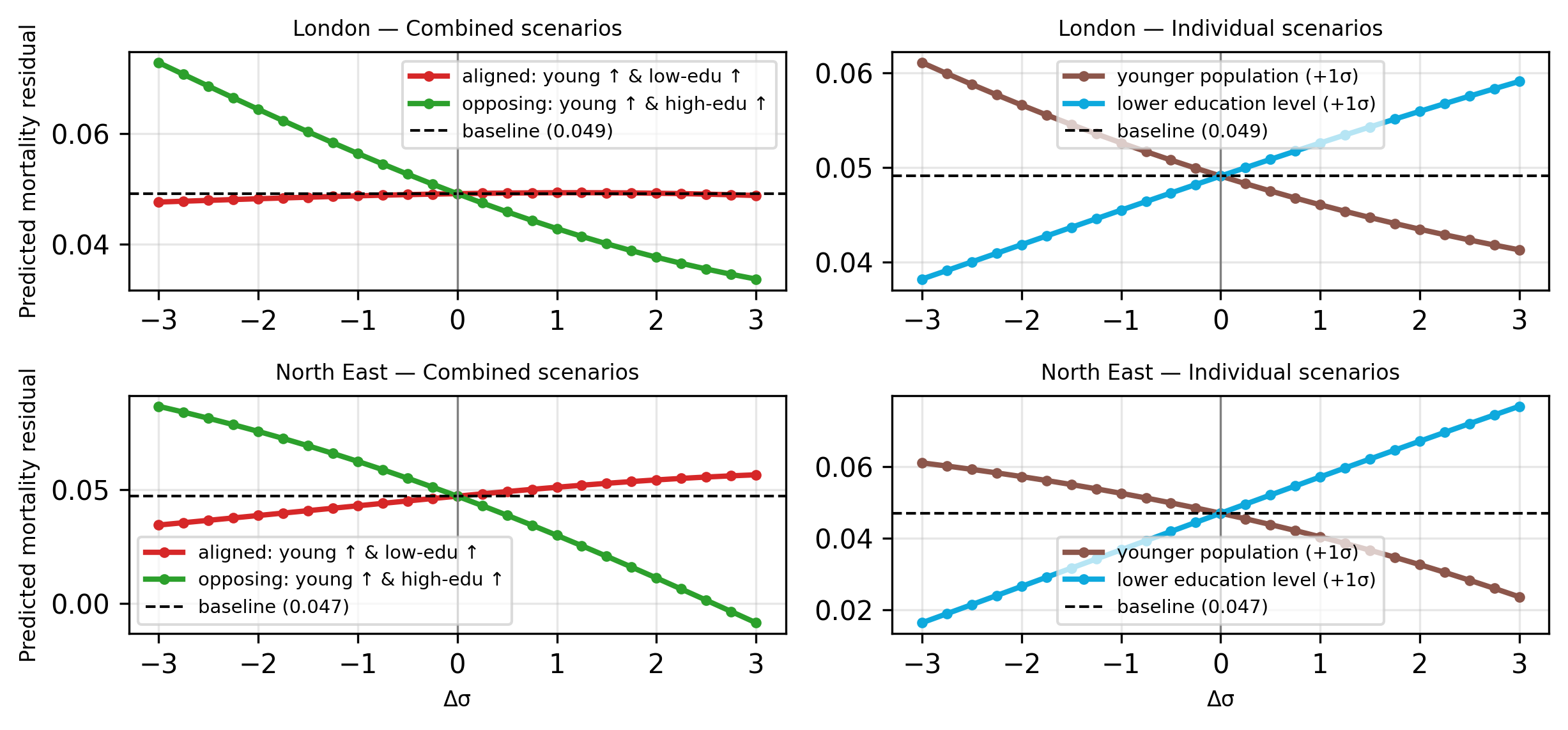}
    \caption{$\Delta$ predictions with varying proportions of young and lower-educated people in OAs, as predicted by the RGNN-OA in London.}
    \label{fig:counterfactual-sweep}
\end{figure}

As an early set of results, we vary census features at OA and examine the changes in predictions made using the RGNN-OA.
We perform two experiments here; increasing the proportion of OA population with low educational qualifications (level 1/2), as well as the proportion of young adults (25-29), by one standard deviation upwards.
We perform this analysis for the London region using temperature data for the 21st July 2022, the hottest day on-record in the UK.

Chloropleth visualisation in Figure~\ref{fig:counterfactual} shows heterogeneity across spatial areas, and a sweep over delta magnitudes in Figure~\ref{fig:counterfactual-sweep} (Appendix) shows the expected relationships.
Increased proportions of population with low educational levels increases mortality, and increased proportions of young people decrease it, in-line with epidemiological literature on deprivation and clinical risks \cite{loCompoundMortalityImpacts2024}.
The North East in-general shows greater sensitivity to demographic changes, particularly in education levels, possibly due to the broadly lower socio-economic status of the region in-comparison to reason.
This spatially-resolved attribution, mapping demographic vulnerability onto urban geography, exemplifies the kind of actionable, area-level intelligence that can directly inform local heat-health interventions.
Further analysis will follow in future iterations of this work.

\subsection{Discussion}

The most consequential result is calibration: where DLNM and RF coverage collapse during the 2022 heatwave, all RGNN variants maintain near-nominal coverage.
For a public health tool intended to support decision-making during the events where it is needed most, this robustness under distributional shift is a meaningful contribution.

Early counterfactual experiments provide evidence that the model is learning epidemiologically coherent structure rather than spurious correlations, with deprivation and age effects in the expected directions. 
More systematic feature attribution analysis will follow in future iterations.
A natural extension is to assess whether the learned demographic embeddings transfer across geographic contexts, enabling risk curve estimation for areas with limited mortality data.

\section{Conclusion}

We introduced Risk Graph Neural Networks, which augment classical DLNMs with a hierarchical GNN encoder over census features, preserving interpretable risk curve outputs while substantially improving predictive calibration under extreme heat.
Evaluated across 10 regions of England and Wales, RGNN variants maintain near-nominal uncertainty coverage where standard baselines collapse.
Future work will broaden the geographic scope, investigate adaptive hierarchy depth, and provide fuller feature attribution analysis.
We also intend to explore whether representations learned over England and Wales generalise to other health outcome data.

\begin{credits}
\subsubsection{\ackname}

\subsubsection{\discintname}
The authors have no competing interests to declare that are relevant
to the content of this article.
\end{credits}

\bibliographystyle{abbrvnat}
\bibliography{references}

\appendix
\section{Appendix}



\begin{longtable}{@{}llccccc@{}}
\caption{%
  Per-region holdout results for all models across both holdout years.
  RMSE$\downarrow$, MAE$\downarrow$, Pearson~$r\,\uparrow$, CRPS$\downarrow$, and Coverage-90
  are shown for each region under the 2018 (moderate) and 2022 (severe heatwave)
  holdout years.
  \textbf{Bold} indicates the best result per metric in each region-year block.
  For Coverage-90, best is defined as closest to the nominal 0.90 level.
  The Poisson noise floor (\textit{fl}) and Pearson ceiling (\textit{cl}) are shown
  for reference alongside each region header.
  $^\dagger$CRPS and Coverage-90 are not reported for LSTM and MLP,
  as these models do not produce well-defined predictive distributions.
\label{tab:supp_perregion}}\\
\toprule
Region & Model & RMSE$\downarrow$ & MAE$\downarrow$ & Pearson$\uparrow$ & CRPS$\downarrow$ & Cov-90 \\
\midrule
\endfirsthead
\multicolumn{7}{l}{\small\textit{(continued from previous page)}}\\
\toprule
Region & Model & RMSE$\downarrow$ & MAE$\downarrow$ & Pearson$\uparrow$ & CRPS$\downarrow$ & Cov-90 \\
\midrule
\endhead
\midrule
\multicolumn{7}{r}{\small\textit{(continued on next page)}}\\
\endfoot
\bottomrule
\multicolumn{7}{@{}p{\linewidth}@{}}{\footnotesize
  \textit{fl}\,=\,Poisson RMSE noise floor;
  \textit{cl}\,=\,Pearson ceiling.
  RGNN\textsubscript{L}\,=\,RGNN-LSOA;
  RGNN\textsubscript{OA}\,=\,RGNN-OA.}\\
\endlastfoot

\multicolumn{7}{@{}l}{\textbf{North East --- 2018}\enspace{\footnotesize\textit{fl}\,=\,0.327,\enspace\textit{cl}\,=\,0.630}}\\
\addlinespace[1pt]
& DLNM & 0.455 & 0.346 & 0.204 & 0.265 & 0.715\\
& RF & 0.384 & 0.286 & 0.405 & 0.251 & 0.342\\
& GBM & \textbf{0.360} & \textbf{0.272} & 0.286 & \textbf{0.203} & 0.808\\
& LSTM$^\dagger$ & 0.372 & 0.296 & \textbf{0.540} & --- & ---\\
& MLP$^\dagger$ & 0.413 & 0.332 & 0.429 & --- & ---\\
& RGNN & 0.409 & 0.341 & 0.266 & 0.236 & \textbf{0.962}\\
& RGNN\textsubscript{L} & 0.378 & 0.306 & 0.432 & 0.220 & 0.984\\
& RGNN\textsubscript{OA} & 0.366 & 0.298 & 0.426 & 0.212 & 0.970\\
\addlinespace[3pt]
\multicolumn{7}{@{}l}{\textbf{North East --- 2022}\enspace{\footnotesize\textit{fl}\,=\,0.326,\enspace\textit{cl}\,=\,0.631}}\\
\addlinespace[1pt]
& DLNM & 0.653 & 0.521 & 0.161 & 0.420 & 0.504\\
& RF & 0.594 & 0.482 & \textbf{0.287} & 0.445 & 0.148\\
& GBM & 0.594 & 0.482 & 0.221 & 0.382 & 0.506\\
& LSTM$^\dagger$ & 0.527 & 0.449 & 0.240 & --- & ---\\
& MLP$^\dagger$ & 0.654 & 0.534 & 0.179 & --- & ---\\
& RGNN & \textbf{0.362} & \textbf{0.260} & 0.255 & \textbf{0.202} & \textbf{0.964}\\
& RGNN\textsubscript{L} & 0.393 & 0.306 & 0.213 & 0.223 & 0.975\\
& RGNN\textsubscript{OA} & 0.379 & 0.285 & 0.247 & 0.211 & \textbf{0.964}\\
\addlinespace[3pt]
\midrule

\multicolumn{7}{@{}l}{\textbf{North West --- 2018}\enspace{\footnotesize\textit{fl}\,=\,0.188,\enspace\textit{cl}\,=\,0.798}}\\
\addlinespace[1pt]
& DLNM & 0.321 & 0.245 & 0.258 & 0.187 & 0.737\\
& RF & 0.287 & 0.211 & 0.471 & 0.185 & 0.351\\
& GBM & 0.271 & 0.205 & 0.303 & 0.155 & 0.758\\
& LSTM$^\dagger$ & 0.249 & 0.194 & \textbf{0.612} & --- & ---\\
& MLP$^\dagger$ & 0.275 & 0.212 & 0.362 & --- & ---\\
& RGNN & 0.291 & 0.242 & 0.459 & 0.171 & 0.997\\
& RGNN\textsubscript{L} & \textbf{0.228} & \textbf{0.180} & 0.566 & 0.140 & 0.997\\
& RGNN\textsubscript{OA} & 0.230 & 0.183 & 0.558 & \textbf{0.139} & \textbf{0.989}\\
\addlinespace[3pt]
\multicolumn{7}{@{}l}{\textbf{North West --- 2022}\enspace{\footnotesize\textit{fl}\,=\,0.187,\enspace\textit{cl}\,=\,0.801}}\\
\addlinespace[1pt]
& DLNM & 0.587 & 0.491 & 0.484 & 0.410 & 0.307\\
& RF & 0.582 & 0.488 & 0.486 & 0.461 & 0.071\\
& GBM & 0.576 & 0.485 & 0.494 & 0.406 & 0.272\\
& LSTM$^\dagger$ & 0.581 & 0.523 & 0.534 & --- & ---\\
& MLP$^\dagger$ & 0.506 & 0.435 & \textbf{0.554} & --- & ---\\
& RGNN & \textbf{0.336} & \textbf{0.248} & 0.485 & \textbf{0.184} & 0.959\\
& RGNN\textsubscript{L} & 0.425 & 0.356 & 0.397 & 0.244 & 0.882\\
& RGNN\textsubscript{OA} & 0.397 & 0.320 & 0.446 & 0.223 & \textbf{0.893}\\
\addlinespace[3pt]
\midrule

\multicolumn{7}{@{}l}{\textbf{Yorkshire --- 2018}\enspace{\footnotesize\textit{fl}\,=\,0.215,\enspace\textit{cl}\,=\,0.743}}\\
\addlinespace[1pt]
& DLNM & 0.330 & 0.242 & 0.262 & 0.189 & 0.721\\
& RF & 0.289 & 0.212 & 0.459 & 0.185 & 0.351\\
& GBM & 0.275 & 0.206 & 0.295 & 0.156 & 0.783\\
& LSTM$^\dagger$ & \textbf{0.239} & \textbf{0.182} & \textbf{0.618} & --- & ---\\
& MLP$^\dagger$ & 0.270 & 0.209 & 0.508 & --- & ---\\
& RGNN & 0.319 & 0.270 & 0.403 & 0.185 & \textbf{0.981}\\
& RGNN\textsubscript{L} & 0.265 & 0.213 & 0.539 & 0.157 & 0.986\\
& RGNN\textsubscript{OA} & 0.262 & 0.210 & 0.521 & \textbf{0.153} & \textbf{0.981}\\
\addlinespace[3pt]
\multicolumn{7}{@{}l}{\textbf{Yorkshire --- 2022}\enspace{\footnotesize\textit{fl}\,=\,0.215,\enspace\textit{cl}\,=\,0.743}}\\
\addlinespace[1pt]
& DLNM & 0.558 & 0.452 & \textbf{0.485} & 0.372 & 0.414\\
& RF & 0.536 & 0.436 & 0.433 & 0.408 & 0.115\\
& GBM & 0.532 & 0.435 & 0.447 & 0.358 & 0.369\\
& LSTM$^\dagger$ & 0.545 & 0.467 & 0.435 & --- & ---\\
& MLP$^\dagger$ & 0.535 & 0.456 & 0.437 & --- & ---\\
& RGNN & \textbf{0.312} & \textbf{0.235} & 0.442 & \textbf{0.175} & 0.953\\
& RGNN\textsubscript{L} & 0.375 & 0.309 & 0.347 & 0.213 & 0.934\\
& RGNN\textsubscript{OA} & 0.351 & 0.278 & 0.404 & 0.196 & \textbf{0.923}\\
\addlinespace[3pt]
\midrule

\multicolumn{7}{@{}l}{\textbf{East Midlands --- 2018}\enspace{\footnotesize\textit{fl}\,=\,0.221,\enspace\textit{cl}\,=\,0.716}}\\
\addlinespace[1pt]
& DLNM & 0.335 & 0.262 & 0.346 & 0.199 & 0.710\\
& RF & 0.290 & 0.218 & 0.425 & 0.190 & 0.392\\
& GBM & 0.283 & 0.212 & 0.335 & 0.162 & 0.742\\
& LSTM$^\dagger$ & 0.240 & 0.188 & \textbf{0.618} & --- & ---\\
& MLP$^\dagger$ & 0.279 & 0.221 & 0.526 & --- & ---\\
& RGNN & 0.256 & 0.210 & 0.503 & 0.150 & \textbf{0.986}\\
& RGNN\textsubscript{L} & 0.226 & 0.184 & 0.551 & 0.136 & 0.989\\
& RGNN\textsubscript{OA} & \textbf{0.225} & \textbf{0.182} & 0.545 & \textbf{0.133} & \textbf{0.986}\\
\addlinespace[3pt]
\multicolumn{7}{@{}l}{\textbf{East Midlands --- 2022}\enspace{\footnotesize\textit{fl}\,=\,0.222,\enspace\textit{cl}\,=\,0.718}}\\
\addlinespace[1pt]
& DLNM & 0.548 & 0.439 & \textbf{0.561} & 0.359 & 0.449\\
& RF & 0.535 & 0.432 & 0.454 & 0.402 & 0.101\\
& GBM & 0.537 & 0.433 & 0.340 & 0.353 & 0.386\\
& LSTM$^\dagger$ & 0.553 & 0.469 & 0.358 & --- & ---\\
& MLP$^\dagger$ & 0.539 & 0.457 & 0.482 & --- & ---\\
& RGNN & \textbf{0.339} & \textbf{0.254} & 0.495 & \textbf{0.184} & 0.945\\
& RGNN\textsubscript{L} & 0.415 & 0.342 & 0.396 & 0.237 & 0.866\\
& RGNN\textsubscript{OA} & 0.389 & 0.310 & 0.449 & 0.217 & \textbf{0.868}\\
\addlinespace[3pt]
\midrule

\multicolumn{7}{@{}l}{\textbf{West Midlands --- 2018}\enspace{\footnotesize\textit{fl}\,=\,0.202,\enspace\textit{cl}\,=\,0.751}}\\
\addlinespace[1pt]
& DLNM & 0.327 & 0.251 & 0.353 & 0.194 & 0.699\\
& RF & 0.291 & 0.218 & 0.529 & 0.192 & 0.345\\
& GBM & 0.288 & 0.215 & 0.426 & 0.165 & 0.764\\
& LSTM$^\dagger$ & 0.253 & 0.205 & \textbf{0.683} & --- & ---\\
& MLP$^\dagger$ & 0.247 & 0.193 & 0.555 & --- & ---\\
& RGNN & 0.238 & 0.193 & 0.519 & 0.144 & \textbf{1.000}\\
& RGNN\textsubscript{L} & 0.214 & 0.173 & 0.562 & 0.133 & \textbf{1.000}\\
& RGNN\textsubscript{OA} & \textbf{0.210} & \textbf{0.169} & 0.557 & \textbf{0.130} & \textbf{1.000}\\
\addlinespace[3pt]
\multicolumn{7}{@{}l}{\textbf{West Midlands --- 2022}\enspace{\footnotesize\textit{fl}\,=\,0.202,\enspace\textit{cl}\,=\,0.751}}\\
\addlinespace[1pt]
& DLNM & 0.554 & 0.457 & \textbf{0.533} & 0.376 & 0.375\\
& RF & 0.541 & 0.450 & 0.530 & 0.420 & 0.068\\
& GBM & 0.539 & 0.447 & 0.469 & 0.369 & 0.325\\
& LSTM$^\dagger$ & 0.559 & 0.497 & 0.460 & --- & ---\\
& MLP$^\dagger$ & 0.481 & 0.405 & 0.527 & --- & ---\\
& RGNN & \textbf{0.330} & \textbf{0.239} & 0.508 & \textbf{0.180} & 0.948\\
& RGNN\textsubscript{L} & 0.397 & 0.325 & 0.433 & 0.224 & 0.904\\
& RGNN\textsubscript{OA} & 0.383 & 0.304 & 0.471 & 0.213 & \textbf{0.901}\\
\addlinespace[3pt]
\midrule

\multicolumn{7}{@{}l}{\textbf{East of England --- 2018}\enspace{\footnotesize\textit{fl}\,=\,0.191,\enspace\textit{cl}\,=\,0.753}}\\
\addlinespace[1pt]
& DLNM & 0.319 & 0.238 & 0.348 & 0.182 & 0.742\\
& RF & 0.285 & 0.205 & 0.407 & 0.179 & 0.378\\
& GBM & 0.265 & 0.194 & 0.395 & 0.147 & 0.769\\
& LSTM$^\dagger$ & 0.281 & 0.216 & 0.494 & --- & ---\\
& MLP$^\dagger$ & 0.261 & 0.199 & 0.508 & --- & ---\\
& RGNN & \textbf{0.215} & \textbf{0.164} & 0.579 & \textbf{0.126} & 0.973\\
& RGNN\textsubscript{L} & 0.220 & 0.172 & \textbf{0.597} & 0.130 & 0.973\\
& RGNN\textsubscript{OA} & 0.220 & 0.168 & 0.590 & 0.127 & \textbf{0.964}\\
\addlinespace[3pt]
\multicolumn{7}{@{}l}{\textbf{East of England --- 2022}\enspace{\footnotesize\textit{fl}\,=\,0.191,\enspace\textit{cl}\,=\,0.756}}\\
\addlinespace[1pt]
& DLNM & 0.445 & 0.359 & \textbf{0.571} & 0.290 & 0.496\\
& RF & 0.438 & 0.354 & 0.476 & 0.326 & 0.153\\
& GBM & 0.442 & 0.355 & 0.327 & 0.288 & 0.447\\
& LSTM$^\dagger$ & 0.450 & 0.388 & 0.365 & --- & ---\\
& MLP$^\dagger$ & 0.412 & 0.358 & 0.383 & --- & ---\\
& RGNN & \textbf{0.327} & \textbf{0.269} & 0.407 & \textbf{0.186} & \textbf{0.901}\\
& RGNN\textsubscript{L} & 0.398 & 0.342 & 0.324 & 0.234 & 0.827\\
& RGNN\textsubscript{OA} & 0.383 & 0.325 & 0.373 & 0.224 & 0.816\\
\addlinespace[3pt]
\midrule

\multicolumn{7}{@{}l}{\textbf{London --- 2018}\enspace{\footnotesize\textit{fl}\,=\,0.135,\enspace\textit{cl}\,=\,0.801}}\\
\addlinespace[1pt]
& DLNM & 0.201 & 0.159 & 0.375 & 0.119 & 0.764\\
& RF & 0.182 & 0.142 & 0.454 & 0.123 & 0.367\\
& GBM & \textbf{0.173} & \textbf{0.136} & 0.427 & \textbf{0.101} & 0.797\\
& LSTM$^\dagger$ & 0.200 & 0.165 & 0.442 & --- & ---\\
& MLP$^\dagger$ & 0.183 & 0.145 & 0.525 & --- & ---\\
& RGNN & 0.180 & 0.150 & \textbf{0.633} & 0.114 & \textbf{1.000}\\
& RGNN\textsubscript{L} & 0.225 & 0.192 & 0.625 & 0.136 & \textbf{1.000}\\
& RGNN\textsubscript{OA} & 0.190 & 0.160 & 0.629 & 0.118 & \textbf{1.000}\\
\addlinespace[3pt]
\multicolumn{7}{@{}l}{\textbf{London --- 2022}\enspace{\footnotesize\textit{fl}\,=\,0.134,\enspace\textit{cl}\,=\,0.819}}\\
\addlinespace[1pt]
& DLNM & 0.352 & 0.299 & 0.450 & 0.243 & 0.392\\
& RF & 0.347 & 0.301 & 0.459 & 0.279 & 0.066\\
& GBM & 0.343 & \textbf{0.294} & 0.363 & 0.239 & 0.350\\
& LSTM$^\dagger$ & 0.351 & 0.320 & 0.466 & --- & ---\\
& MLP$^\dagger$ & \textbf{0.334} & 0.299 & \textbf{0.599} & --- & ---\\
& RGNN & 0.366 & 0.328 & 0.471 & \textbf{0.218} & \textbf{0.901}\\
& RGNN\textsubscript{L} & 0.420 & 0.381 & 0.448 & 0.254 & 0.847\\
& RGNN\textsubscript{OA} & 0.379 & 0.341 & 0.468 & 0.227 & 0.874\\
\addlinespace[3pt]
\midrule

\multicolumn{7}{@{}l}{\textbf{South East --- 2018}\enspace{\footnotesize\textit{fl}\,=\,0.157,\enspace\textit{cl}\,=\,0.807}}\\
\addlinespace[1pt]
& DLNM & 0.272 & 0.205 & 0.315 & 0.156 & 0.770\\
& RF & 0.246 & 0.181 & 0.383 & 0.156 & 0.359\\
& GBM & 0.232 & 0.172 & 0.370 & 0.130 & 0.806\\
& LSTM$^\dagger$ & 0.275 & 0.199 & 0.300 & --- & ---\\
& MLP$^\dagger$ & 0.252 & 0.200 & 0.442 & --- & ---\\
& RGNN & \textbf{0.198} & \textbf{0.154} & 0.519 & \textbf{0.119} & 0.986\\
& RGNN\textsubscript{L} & 0.202 & 0.161 & \textbf{0.530} & 0.121 & 0.984\\
& RGNN\textsubscript{OA} & 0.203 & 0.159 & 0.526 & 0.119 & \textbf{0.975}\\
\addlinespace[3pt]
\multicolumn{7}{@{}l}{\textbf{South East --- 2022}\enspace{\footnotesize\textit{fl}\,=\,0.157,\enspace\textit{cl}\,=\,0.808}}\\
\addlinespace[1pt]
& DLNM & 0.466 & 0.382 & \textbf{0.569} & 0.314 & 0.419\\
& RF & 0.464 & 0.380 & 0.462 & 0.354 & 0.126\\
& GBM & 0.465 & 0.381 & 0.430 & 0.316 & 0.356\\
& LSTM$^\dagger$ & 0.474 & 0.414 & 0.426 & --- & ---\\
& MLP$^\dagger$ & 0.449 & 0.393 & 0.510 & --- & ---\\
& RGNN & \textbf{0.340} & \textbf{0.282} & 0.442 & \textbf{0.194} & \textbf{0.888}\\
& RGNN\textsubscript{L} & 0.407 & 0.351 & 0.394 & 0.242 & 0.792\\
& RGNN\textsubscript{OA} & 0.412 & 0.354 & 0.418 & 0.247 & 0.759\\
\addlinespace[3pt]
\midrule

\multicolumn{7}{@{}l}{\textbf{South West --- 2018}\enspace{\footnotesize\textit{fl}\,=\,0.213,\enspace\textit{cl}\,=\,0.740}}\\
\addlinespace[1pt]
& DLNM & 0.316 & 0.239 & 0.319 & 0.180 & 0.770\\
& RF & 0.279 & 0.205 & 0.469 & 0.177 & 0.375\\
& GBM & 0.263 & 0.195 & 0.439 & 0.147 & \textbf{0.833}\\
& LSTM$^\dagger$ & 0.266 & 0.193 & 0.518 & --- & ---\\
& MLP$^\dagger$ & 0.258 & 0.191 & 0.510 & --- & ---\\
& RGNN & 0.244 & 0.196 & 0.543 & 0.141 & 0.970\\
& RGNN\textsubscript{L} & 0.231 & 0.178 & \textbf{0.580} & 0.134 & 0.984\\
& RGNN\textsubscript{OA} & \textbf{0.220} & \textbf{0.170} & 0.565 & \textbf{0.127} & 0.981\\
\addlinespace[3pt]
\multicolumn{7}{@{}l}{\textbf{South West --- 2022}\enspace{\footnotesize\textit{fl}\,=\,0.214,\enspace\textit{cl}\,=\,0.741}}\\
\addlinespace[1pt]
& DLNM & 0.570 & 0.462 & \textbf{0.544} & 0.378 & 0.430\\
& RF & 0.557 & 0.456 & 0.468 & 0.426 & 0.090\\
& GBM & 0.560 & 0.460 & 0.394 & 0.377 & 0.361\\
& LSTM$^\dagger$ & 0.558 & 0.487 & 0.448 & --- & ---\\
& MLP$^\dagger$ & 0.509 & 0.439 & 0.511 & --- & ---\\
& RGNN & \textbf{0.379} & \textbf{0.294} & 0.524 & \textbf{0.209} & \textbf{0.866}\\
& RGNN\textsubscript{L} & 0.521 & 0.447 & 0.467 & 0.323 & 0.630\\
& RGNN\textsubscript{OA} & 0.441 & 0.362 & 0.505 & 0.259 & 0.745\\
\addlinespace[3pt]
\midrule

\multicolumn{7}{@{}l}{\textbf{Wales --- 2018}\enspace{\footnotesize\textit{fl}\,=\,0.303,\enspace\textit{cl}\,=\,0.651}}\\
\addlinespace[1pt]
& DLNM & 0.451 & 0.345 & 0.376 & 0.265 & 0.701\\
& RF & 0.386 & 0.291 & 0.474 & 0.257 & 0.321\\
& GBM & 0.380 & 0.288 & 0.411 & 0.219 & 0.756\\
& LSTM$^\dagger$ & 0.340 & 0.271 & 0.611 & --- & ---\\
& MLP$^\dagger$ & 0.361 & 0.295 & 0.615 & --- & ---\\
& RGNN & 0.310 & 0.246 & 0.534 & 0.179 & \textbf{0.973}\\
& RGNN\textsubscript{L} & \textbf{0.272} & \textbf{0.216} & \textbf{0.616} & \textbf{0.160} & 0.989\\
& RGNN\textsubscript{OA} & 0.279 & 0.219 & 0.610 & 0.162 & 0.981\\
\addlinespace[3pt]
\multicolumn{7}{@{}l}{\textbf{Wales --- 2022}\enspace{\footnotesize\textit{fl}\,=\,0.302,\enspace\textit{cl}\,=\,0.651}}\\
\addlinespace[1pt]
& DLNM & 0.637 & 0.504 & 0.502 & 0.405 & 0.496\\
& RF & 0.612 & 0.483 & \textbf{0.543} & 0.447 & 0.142\\
& GBM & 0.607 & 0.480 & 0.490 & 0.385 & 0.461\\
& LSTM$^\dagger$ & 0.610 & 0.546 & 0.475 & --- & ---\\
& MLP$^\dagger$ & 0.619 & 0.545 & 0.488 & --- & ---\\
& RGNN & \textbf{0.412} & \textbf{0.302} & 0.499 & \textbf{0.219} & 0.926\\
& RGNN\textsubscript{L} & 0.486 & 0.399 & 0.413 & 0.277 & 0.868\\
& RGNN\textsubscript{OA} & 0.456 & 0.358 & 0.468 & 0.252 & \textbf{0.882}\\
\addlinespace[3pt]

\end{longtable}

\begin{figure}[h]
    \centering
    \includegraphics[width=\linewidth]{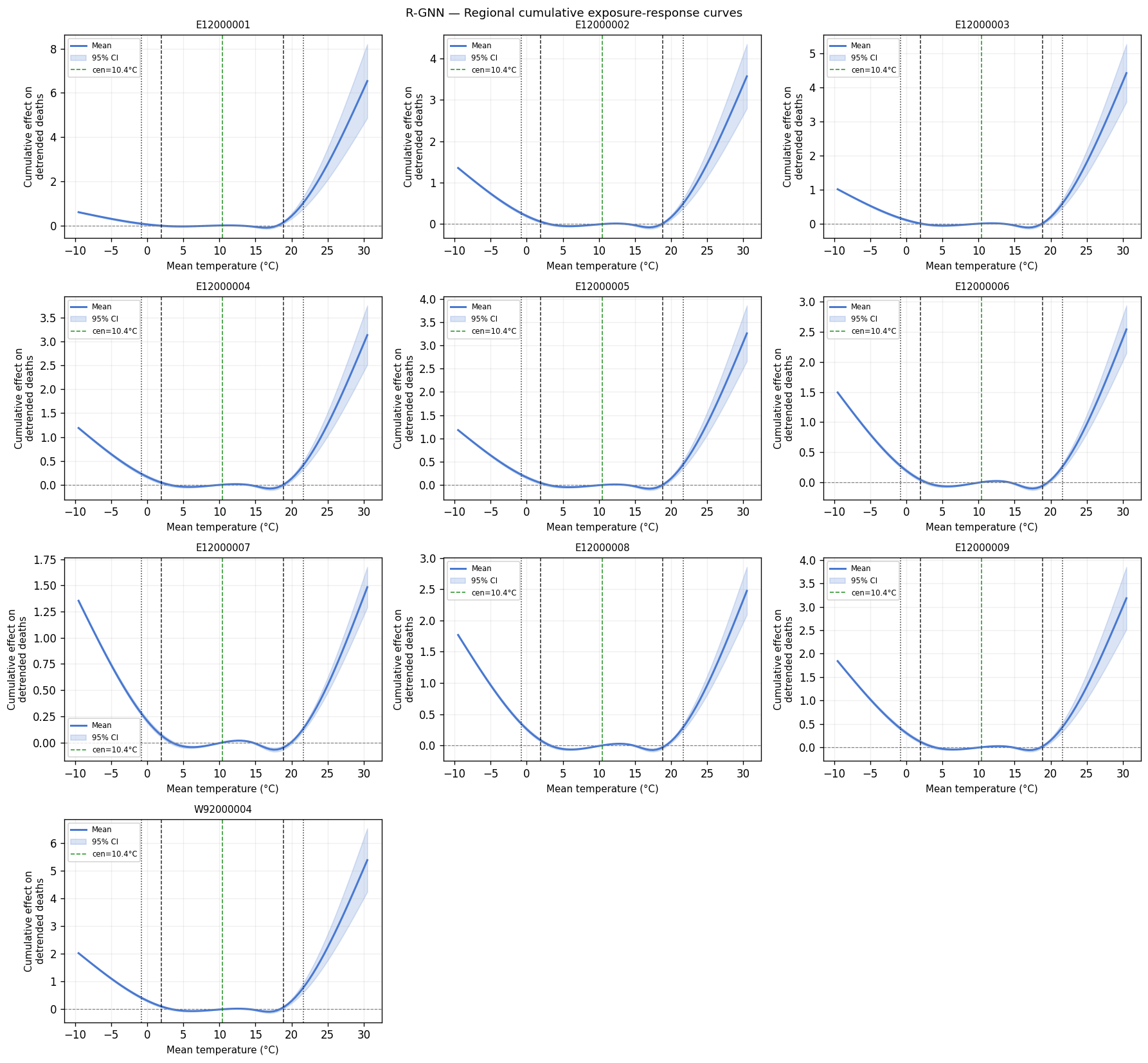}
    \caption{Vanilla DLNM curves for each UK region in our experiments. Important to note in contrast with the optimised curves below is the flat portion at moderate temperatures, with a `bump' due to the combination of spline specification and heat risk falling below detectable SNR levels.}
    \label{fig:curve-init}
\end{figure}

\begin{figure}[h]
    \centering
    \includegraphics[width=\linewidth]{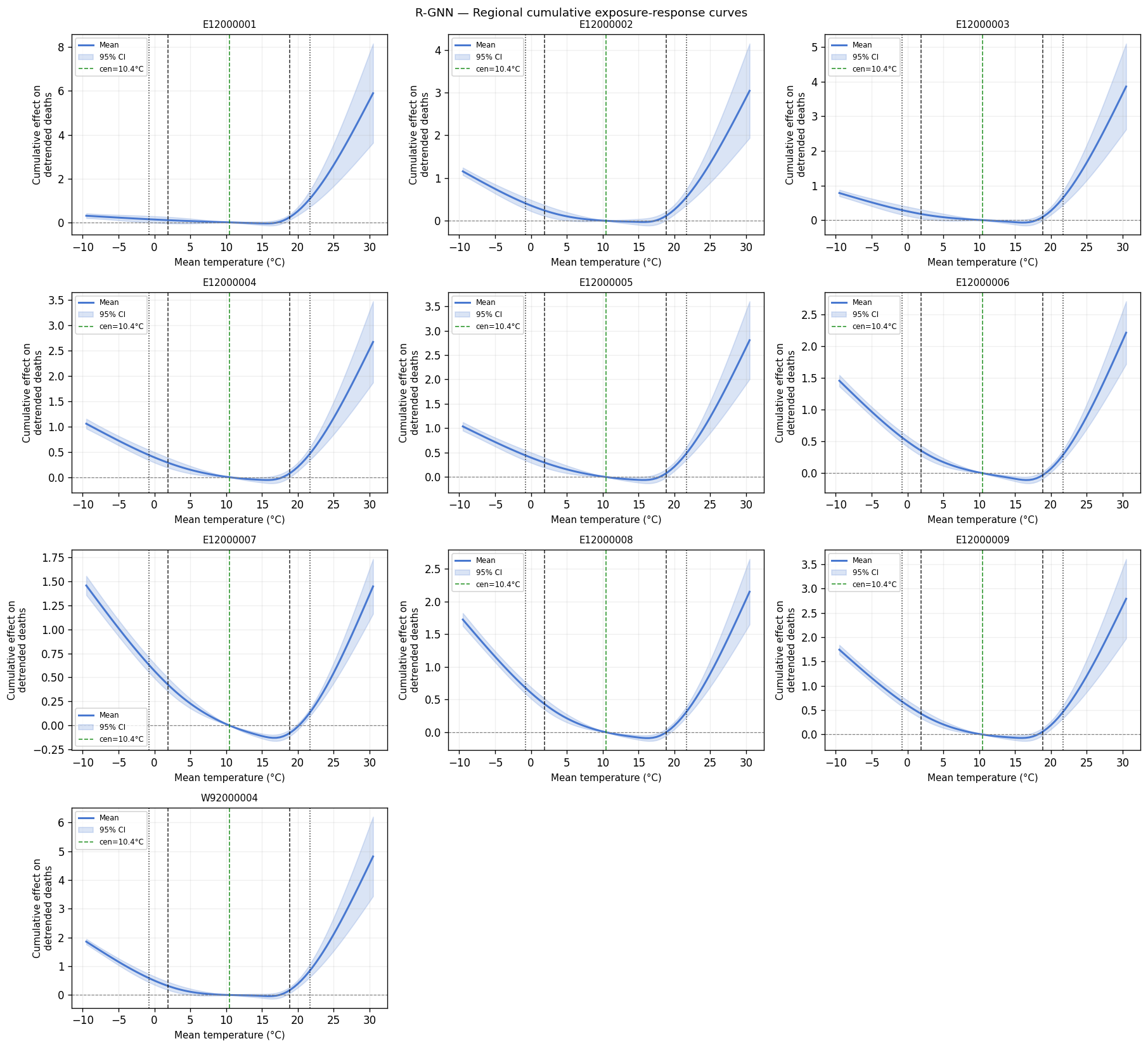}
    \caption{DLNM curves for each UK region in our experiments, optimised by RGNN-LAD.}
    \label{fig:curve-lad}
\end{figure}

\begin{figure}[h]
    \centering
    \includegraphics[width=\linewidth]{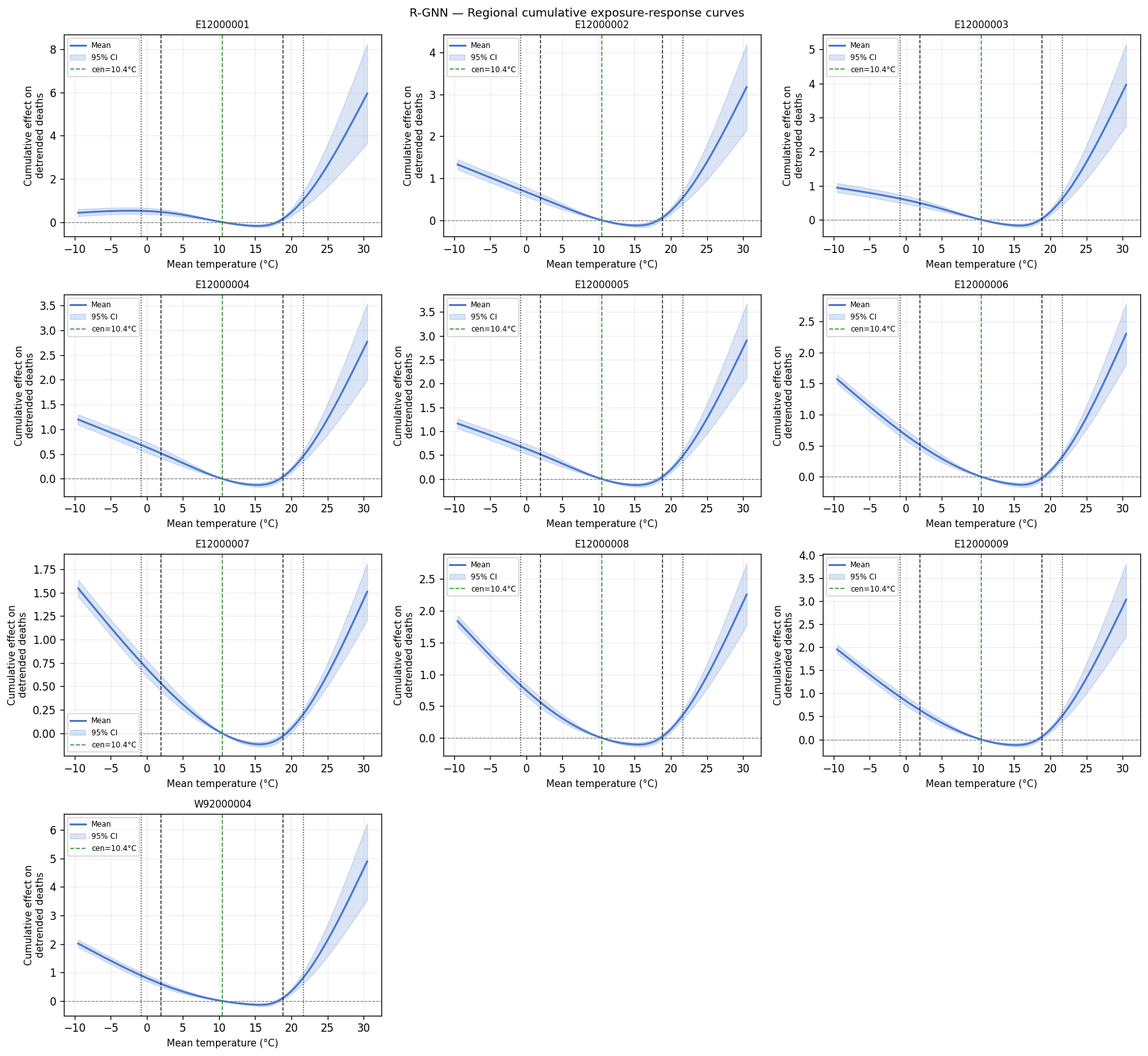}
    \caption{DLNM curves for each UK region in our experiments, optimised by RGNN-LSOA.}
    \label{fig:curve-lsoa}
\end{figure}

\begin{figure}[h]
    \centering
    \includegraphics[width=\linewidth]{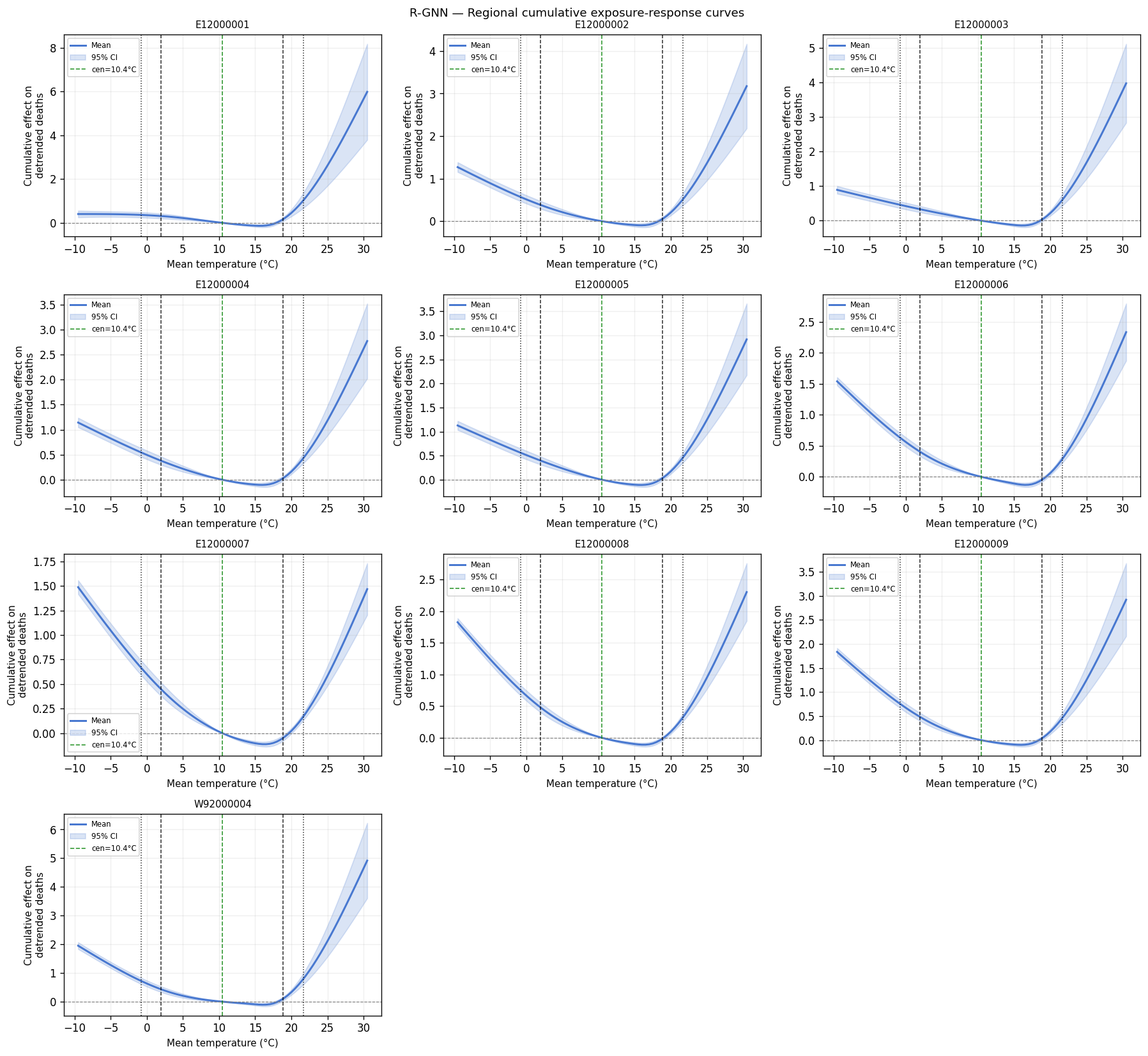}
    \caption{DLNM curves for each UK region in our experiments, optimised by RGNN-OA.}
    \label{fig:curve-oa}
\end{figure}

\end{document}